\documentclass{ifacconf}
\usepackage{graphicx}      
\usepackage{subcaption}
\usepackage{amsmath}
\usepackage{natbib}        
\usepackage{placeins}

\DeclareMathOperator*{\argmin}{arg\,min}

\begin{document}
\begin{frontmatter}

\title{Fast Expanding Safe Circular Regions for Efficient Local Path Planning} 

\thanks[footnoteinfo]{This work has been partially funded by the European Union’s Horizon Europe Research and Innovation Program, under the Grant Agreement No. 101119774 SPEAR.}

\author{Scott Fredriksson,} 
\author{Akshit Saradagi,} 
\author{George Nikolakopoulos}

\address{Robotics and Artificial Intelligence Team\\Luleå University of Technology, 
   Luleå, Sweden\\ (e-mail: scofre@ltu.se).}

\begin{abstract}
Local navigation is one of the fundamental problems in robot navigation, and numerous approaches have been proposed over the years, including methods such as the Dynamic Window Approach, Model Predictive Control, and more recently, Control Barrier Functions and machine learning–based techniques. While these methods perform well in simple environments, many of them rely on optimization or learning-based procedures that can struggle in more complex scenarios. 
In contrast, this article proposes a more geometric-algorithmic approach that enables a local navigation method with faster computation times and longer planning horizons. The proposed method is based on the computation of a sequence of circular regions from a local LiDAR scan that expand in the direction of the goal and capture free local navigable space. The proposed method was implemented in the ROS2 framework and evaluated in a simulated environment.
\end{abstract}


\end{frontmatter}

\section{Introduction}
As one of the fundamental elements of mobile robot autonomy, path planning is critical for enabling autonomous operation, in both map-based and reactive approaches. Traditionally, map-based path planning has been divided into three modules: a global planner, a local planner, and control, \citep{Cybulski2019}. The global planner is responsible for generating an overarching path from the robot’s current position to the goal pose. However, this global path can be computationally expensive to update when using optimal planners \citep{Duchon2014} or potentially unsafe and suboptimal when relying on sampling-based planners \citep{Noreen2016}. Moreover, global planners or mapping solutions sometimes neglect or miss smaller obstacles in the environment. To address these limitations, a faster and more responsive local planner is typically employed. The purpose of the local planner is to compute a safe, short-term local path that adheres to the global plan while respecting the robot’s dynamic and kinematic constraints.

The most well-known local planner is the potential field method and its variants \citep{Khatib1986}. While it is one of the simplest local planners, it suffers from local minima, causing the robot to become easily trapped. Most local planners operate within a finite horizon, where they attempt to find a feasible local path. As long as a valid solution exists within this horizon, a good planner should be able to identify it. One of the earliest implementations of this idea is the Dynamic Window Approach (DWA) \citep{Fox1997}. Another strategy is to modify the existing global path \citep{Quinlan1993}. Methods of this type are used in planners such as the Timed Elastic Bands (TEB) local planner \citep{Roesmann2015,Chen2025a}.

The local path finding problem has also been investigated from a control-theoretic perspective. By incorporating obstacle representations into Model Predictive Control (MPC), a safe control velocity can be generated for navigation \citep{Brito2019,Liu2024}. More recently, control barrier functions \citep{Ames2017} have been applied to ensure safe local path generation \citep{Long2021,Dai2023,Saradagi2024}.

Machine learning has also been explored for solving the local path planning problem \citep{wang2023deep,MartinezBaselga2025}, using reinforcement learning, while \cite{Saucedo2023a} uses vision-based approaches. Hybrid methods that combine learning with control-theoretic techniques have also been proposed \citep{Long2021}.

The main drawback of the current state-of-the-art methods is their reliance on computationally intensive, optimization or learning-based approaches. These methods, limited to account for a small number of obstacles, often result in slower computation, shorter prediction horizons, increased risk of becoming trapped in local minima, encountering deadlocks, or facing ill-conditioned optimization problems. 

\begin{figure*}[t]
    \centering
    \includegraphics[width=1\linewidth]{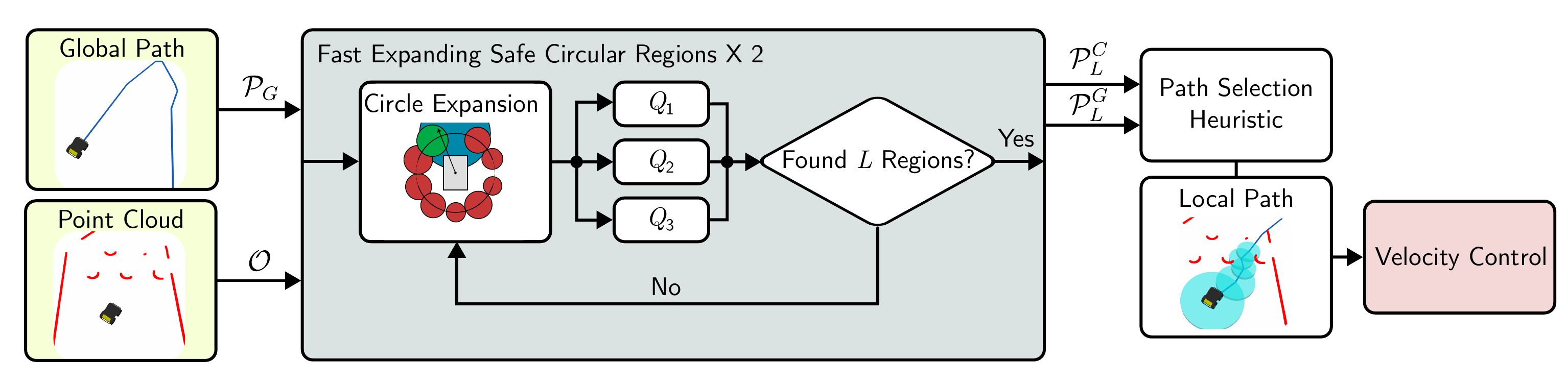}
    \caption{Overview of the proposed geometric-algorithmic approach for safe local navigation.}
    \label{fig:over}
\end{figure*}

\section{Contribution}
To address this problem, this article proposes the method, Fast Expanding Safe Circular Regions (FESCR), an algorithmic geometric approach tailored for modern robotic ground platforms that can turn on the spot. The method supports ground robots that are holonomic, or capable of turning in place, relying solely on data from the current LiDAR scan. It is designed to maintain a safe and comfortable distance from obstacles while being able to navigate through passages as narrow as the robot itself when necessary. 

The proposed FESCR is combined with a simple velocity controller, resulting in a robust and computationally efficient local planner. Figure~\ref{fig:over} shows an overview of the proposed method. The proposed approach was evaluated in a simulated environment\footnote{Video of simulation results: \\https://www.youtube.com/watch?v=RnOdL5bxskw} using different configurations of the proposed method.

\section{Problem Formulation} \label{sec:prob}
Consider a robot with a rectangular footprint, with a width $R_{\text{width}}$ and length $R_{\text{length}}$, that operates within a local environment. The objective is to determine a local path $\mathcal{P}_L$ that follows the general direction of the global path $\mathcal{P}_G$ while avoiding obstacles using the local LiDAR point cloud. Let $\mathcal{O} = \{o_1, \dots, o_N\}$ denote the set of obstacle points in the point cloud, where each $o_i$ corresponds to a point on the local 2D ground plane surrounding the robot.

To construct the local path, the methodology proposed in this article computes a sequence of $L$ obstacle-free circular regions $\mathcal{C} = \{c_1, \dots, c_L\}$. Each circle $c_l = \{c_{lr}, c_{lp}\}$ is defined by its center point $c_{lp}$ and radius $c_{lr}$. The constant $L$ is a user-tunable parameter.

\section{Methodology}
In FESCR, the circular regions are arranged such that the center of each circle lies on the boundary of the preceding circle, except for the first circle, whose center is the same as the robot's center.
Since the center point of the first circular region is fixed, its radius is determined by the distance to the nearest obstacle point, as in Eq.~\eqref{eq:circle_radius}, where $||\bullet||$ denotes the Euclidean norm.
\begin{equation}
    \label{eq:circle_radius}
    c_{lr} = \min_{o \in \mathcal{O}} \left( ||o - c_{lp}|| \right)
\end{equation}
If the first circle exceeds the upper limit, $R_{\text{comfort}}$, chosen by the user, its radius is reduced to $R_{\text{comfort}}$.

\subsection{Expanding Child Circles} \label{sec:expand}
\begin{figure}[b]
    \centering
    \includegraphics[width=0.7\linewidth]{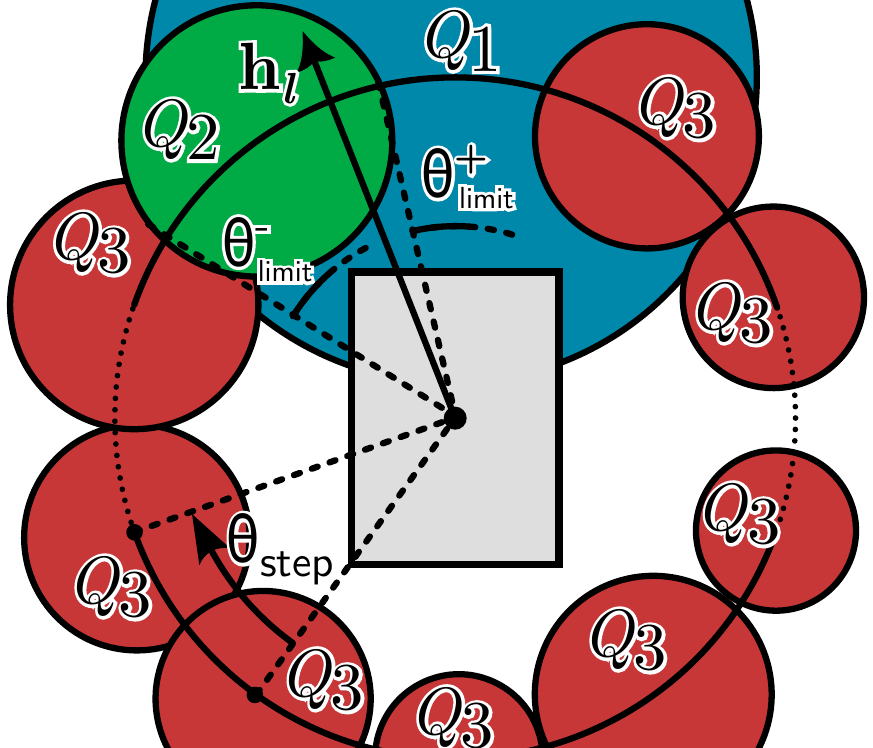}
    \caption{Example of circle expansion: each child region is marked with its respective set $Q_1$, $Q_2$, and $Q_3$. The $\theta_{\text{limit}}^\pm$ is shown for the child circle that belong to $Q_2$.}
    \label{fig:circle_expand}
\end{figure}
When a parent circle is expanded, its child circles are positioned along its boundary at fixed angular intervals, $\theta_{\text{step}}$, specified by the user, as illustrated in Figure~\ref{fig:circle_expand}. Because the center point of each child circle is known, Eq.~\eqref{eq:circle_radius} can then be used to determine its radius.

Each child circular region has its radius constrained between $R_{\text{width}}/2$ and $R_{\text{comfort}}$. If a circular region is smaller than this range, it is discarded, as it is too small for the robot; if it exceeds the upper limit, its radius is reduced to $R_{\text{comfort}}$.

If the radius of the circular region is smaller than the circular footprint radius of the robot, $R_{\text{circler}}$ defined as:
\begin{equation}
     R_{\text{circler}} = \sqrt{\left(\frac{R_{\text{width}}}{2}\right)^2 + \left(\frac{R_{\text{length}}}{2}\right)^2},
\end{equation}
the robot cannot freely turn on the spot without risk of colliding with an obstacle. Therefore, when searching for child circles, the circle centers are only searched within an angular range of $\pm\theta_{piv}$ in front of and behind the robot, where $\theta_{piv}$ is defined as
\begin{equation}
    \begin{matrix}
    \theta_{piv}=
    \left\{   
    \begin{matrix}
       \pi & \text{if } & c_{lr} > R_{\text{circler}}, \\
        T(c_{lr})& \text{if }& R_{\text{width}}/2 \ge c_{lr} \ge R_{\text{circler}}, \\
       0 & \text{if } & R_{\text{width}}/2 > c_{lr}. \\
    \end{matrix}
    \right.\\
    \text{where} \\
    T(c_{lr})=\frac{\pi}{2} - \cos^{-1}\left(\frac{2\cdot c_{lr} - R_{\text{width}}}{2 \cdot R_{\text{circler}} - R_{\text{width}}}\right) 
    \end{matrix}
\end{equation}
This will limit the search so that the children of the parent circle are reachable by the robot even when it is limited in its turning capability.

When the reachable angle is smaller than a full circle and a region behind the robot is selected, the robot will move backward.

\subsection{Selecting Child Circle} \label{sec:select}
The proposed method employs a greedy search strategy and expands nodes only when necessary and will return the first local path that contains $L$ number of circular regions, where none of the center points is within another circular region. Given a local heading direction $\mathbf{h}_{l}$, all child circles, $c_{(l+1)}$, of circle $c_{l}$ are sorted into three distinct sets: $Q_{1}$, consisting of circles with a radius of $R_{\text{comfort}}$ that overlaps the heading direction; $Q_{2}$, containing circles smaller than $R_{\text{comfort}}$ that also overlap the heading direction; and $Q_{3}$, including circles that do not overlap the heading direction. 
To check whether a child circle overlaps the heading direction $\mathbf{h}_{l}$, the condition in Equation~\eqref{eq:overlap_con} is used, where $h_\theta$ is the angle of the heading direction, $\theta_{\text{limit}}^\pm$ are the lower and upper bounds of the allowed angle interval, and $c_{(l+1)\theta}$ is the angle of the vector from $c_{l}$ to $c_{(l+1)}$. If the condition is satisfied, the child circle is considered to overlap the heading direction.

\begin{equation}
\begin{matrix}
     \theta_{\text{limit}}^- \leq h_\theta \leq \theta_{\text{limit}}^+ \vspace{3px} \\
     \text{where} \\ \vspace{3px}
     \theta_{\text{limit}}^\pm = c_{(l+1)\theta} \pm \tan^{-1}\left(\frac{c_{(l+1)r}}{c_{lr}}\right)
\end{matrix}
\label{eq:overlap_con}
\end{equation}

Each time the search is expanded, a child circular region $c_{l+1}$ is selected from the first nonempty set among $Q_{1}$, $Q_{2}$, and $Q_{3}$, in this order. Each set employs its own selection criterion for choosing a node within the set:

\begin{itemize}
    \item \textbf{$Q_{1}$:} The child circle is selected based on the \textit{minimum angular deviation} between the vector from the parent circle center $c_{lp}$ to the child circle center $c_{(l+1)p}$ and the heading vector $\mathbf{h}_l$. The selection criterion is defined as
    \begin{equation}
        \label{eq:q1_criterion}
        c_{l+1}^{*} = \arg\min_{c_{l+1} \in Q_{1}} \, \theta = \arccos\!\left( \frac{(\mathbf{c}_{(l+1)p} - \mathbf{c}_{lp}) \cdot \mathbf{h}_l}{\| \mathbf{c}_{(l+1)p} - \mathbf{c}_{lp} \| \, \| \mathbf{h}_l \|} \right),
    \end{equation}
    where $\theta$ is the angle between the two vectors.

    \item \textbf{$Q_{2}$:} The child circle with the \textit{largest radius} is selected:
    \begin{equation}
        c_{l+1}^{*} = \arg\max_{c_{l+1} \in Q_{2}} \, c_{(l+1)r},
    \end{equation}

    \item \textbf{$Q_{3}$:} The child circle is again selected based on the \textit{minimum angular deviation} from the heading direction, using the same criterion as in Equation~\eqref{eq:q1_criterion}.
\end{itemize}

The motivation behind this approach is that by prioritizing the largest circle, even if it is not the closest to the heading direction, the distance between the robot and the nearest obstacle is maximized, up to a distance of $R_{\text{comfort}}$, while still allowing the robot to follow the intended path.

If $Q_{1}$, $Q_{2}$, and $Q_{3}$ are empty. The parent circle is removed, and the search is continued by selecting a new child from the previous parent circle.

The process described in Subsections \ref{sec:expand} and \ref{sec:select} is repeated until a series of $L$ circular regions is obtained, as described in Section \ref{sec:prob} and illustrated in Figure~\ref{fig:over}.

\subsection{Selecting Heading and Path}
The heading vector, $\mathbf{h}_l$, used in Subsection \ref{sec:select} is derived from the previous set of local paths such that it points from the current circle region to the circle region that was two steps ahead in the previous local path, i.e.,
\begin{equation}
    \mathbf{h}_l = c^{-}_{(l+2)p} - c_{lp},
\end{equation}
where $c^{-}$ denotes a circle region from the previous path.

The heading vector of the last circle is determined from the path by selecting the first point that lies farther from the robot than the circle’s region length. Specifically:
\begin{equation}
\begin{matrix}
    \mathbf{h}_l = \mathbf{p}^\star - c_{lp}, \\
    \text{where} \\
    \mathbf{p}^\star = \min_{\mathbf{p}_i \in \mathcal{P}_G} \left\{ \mathbf{p}_i \,\middle|\, 
    \left\| \mathbf{p}_i - \mathbf{R_{pos}} \right\| \ge 
    \left\| \mathbf{R_{pos}} - c_{lp} \right\| + c_{lr} \right\}.
\end{matrix}
\label{eq:heading}
\end{equation}

This approach yields a consistent local path; however, it may become trapped in local minima when obstacles appear and disappear in the LiDAR view of the robot. To mitigate this issue, an additional search is performed using only the heading derived from \eqref{eq:heading}, producing two local paths, $\{\mathcal{P}_L^C, \mathcal{P}_L^G\}$. The idea is to use the consistent path ($\mathcal{P}_L^C$) under normal conditions and switch to the greedier path ($\mathcal{P}_L^G$) when necessary. The selected path is determined by the following heuristic:
\begin{equation}
    \argmin_{\mathcal{P}_L \in \{\mathcal{P}_L^C, \mathcal{P}_L^G\}} 
    \left( 
         \|\mathcal{P}_L\| + \|\mathcal{P}_L^{end} - p^*\| \right)\, \delta(\mathcal{P}_L)
        + \|\mathcal{P}_G(p^*, \text{end})\|
\end{equation}
where $\|\mathcal{P}_L\|$ is the local path length and $\|\mathcal{P}_G(p^*, \text{end})\|$ is the global path length from point $p^*$ to the end of the global path. The value $\delta(\mathcal{P}_L)$ is defined as:
\begin{equation}
    \delta(\mathcal{P}_L) =
    \begin{cases}
        w_p, & \text{if } \mathcal{P}_L = \mathcal{P}_L^C \\
        1, & \text{if } \mathcal{P}_L = \mathcal{P}_L^G
    \end{cases}
\end{equation}
where $w_p$ is a tunable weight that allows the heuristic to favor the consistent path until it becomes suboptimal, resulting in smoother behavior while still enabling timely switching to the greedier path when necessary.

\section{Velocity Control}
The sequence of circular regions generated by the proposed method can be used with a wide variety of control schemes or as a safety filter in combination with another local planner, such as a machine learning–based local planner. In this article, a very simple velocity controller is used, relying only on the center of the first child circular region, as well as the size of the first region, to determine the robot’s input velocities.

The velocity control scheme is defined using two components: a yaw controller and a forward velocity selector. Let $\theta$ denote the robot's current heading, and let $\theta_d$ be the desired heading toward the center of the first child region. 
The yaw error, $e_\theta$, is defined as the shortest signed angular distance between the current heading $\theta$ and the desired heading $\theta_d$.

The yaw rate is generated using a proportional controller,
\begin{equation}
\omega = k_p\, e_\theta,
\end{equation}
and the forward velocity $v$ follows
\begin{equation}
v = 
\begin{cases}
0, & \text{if } |e_\theta| > \epsilon,\\
v_R, & \text{otherwise},
\end{cases}
\end{equation}
where $\epsilon$ is the yaw-error threshold, and $v_R$ is the robot velocity.

Both the yaw-error threshold $\epsilon$ and the robot speed $v_R$ are adjusted between $\epsilon_\text{min}$ and $\epsilon_\text{max}$, and between $v_\text{min}$ and $v_\text{max}$, respectively, depending on the radius of the first circle region. The adjustments are computed using:

\begin{equation}
    \epsilon = \epsilon_\text{min} + s \cdot (\epsilon_\text{max} - \epsilon_\text{min})
\end{equation}

\begin{equation}
    v_R = v_\text{min} + s\cdot (v_\text{max} - v_\text{min})
\end{equation}

where $s$ is a scaling factor defined as:
\begin{equation}
    s = \frac{2 c_{1r} - R_{\text{width}}}{R_{\text{comfort}} - R_{\text{width}}}.
\end{equation}
\section{Validation} \label{sec:val}
\begin{table}[h]
    \centering
    \begin{tabular}{|c|c||c|c|} \hline
        Variable  & Value  & Variable  & Value \\ \hline
        $R_{\text{width}}$     &  $0.75$ m & $\epsilon_\text{min}$ & $0.2$ rad\\
        $R_{\text{length}}$  & $1.1$ m & $\epsilon_\text{max}$ & $0.4$ rad \\
        $R_{\text{comfort}}$ & $1.5$ m & $\omega_\text{max}$ & $0.8$ rad/s \\
        $v_\text{min}$ & $0.2$ m/s & $\theta_{\text{step}}$ & $0.06$ rad \\
        $v_\text{max}$ & $1.0$ m/s & $k_p$ & $2$ \\
        $w_{p}$ & $0.7$ &  &  \\ \hline
    \end{tabular}
    \caption{Parameters used in simulation.}
    \label{tab:vals}
\end{table}
\begin{figure}[b]
    \centering
    \begin{subfigure}{0.49\linewidth}
    \includegraphics[width=1\linewidth]{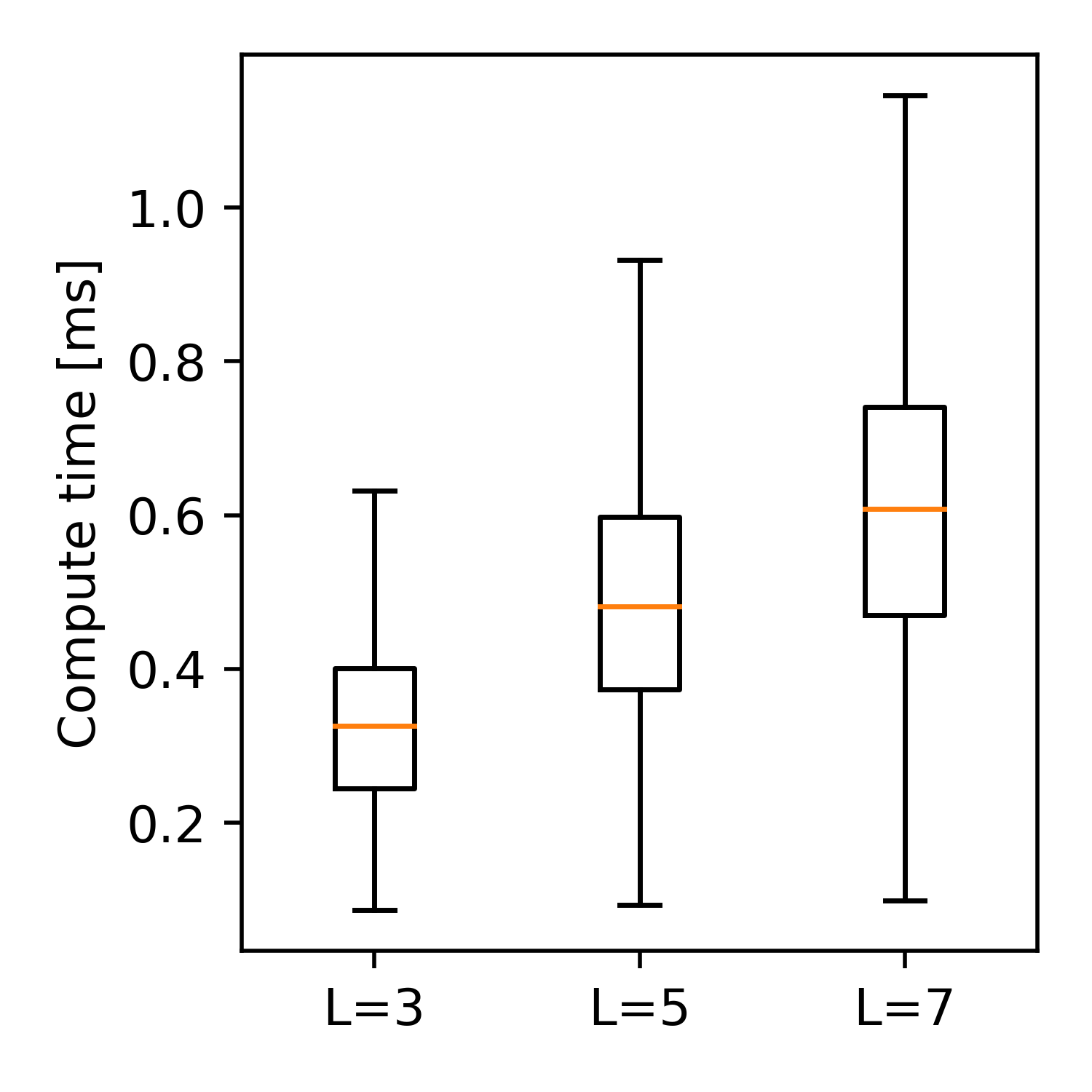}
    \label{fig:res_comput_time1}
    \vspace*{-7mm}
    \caption{Without obstacles.}
    \end{subfigure}
    \begin{subfigure}{0.49\linewidth}
    \includegraphics[width=1\linewidth]{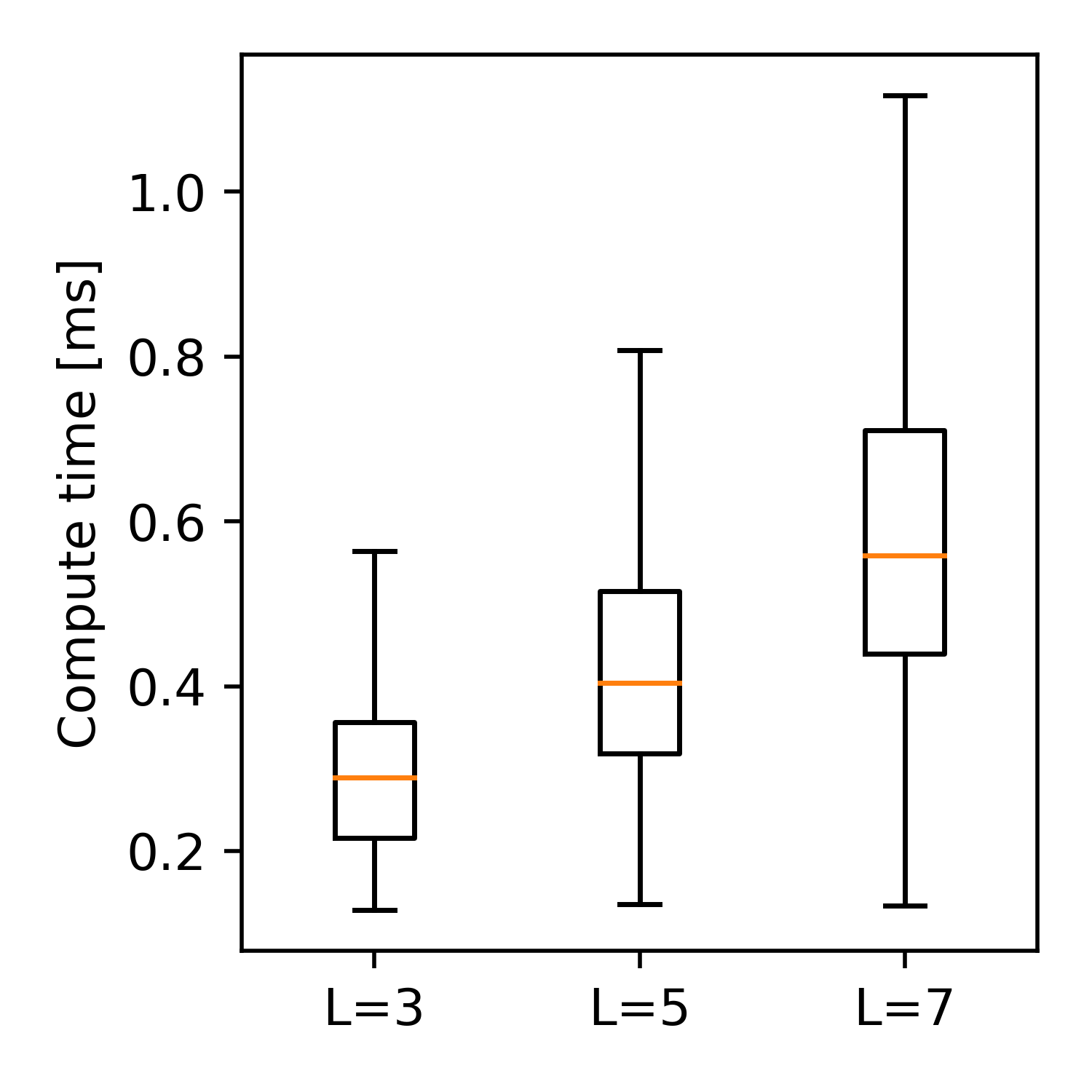}
    \label{fig:res_comput_time2}
    \vspace*{-7mm}
    \caption{With obstacles.}
    \end{subfigure}
    \caption{Computation time for different configurations of $L$.}
    \label{fig:res_comput_time}
\end{figure}
\begin{figure}[t]
    \centering
    \begin{subfigure}{1\linewidth}
    \includegraphics[width=1\linewidth]{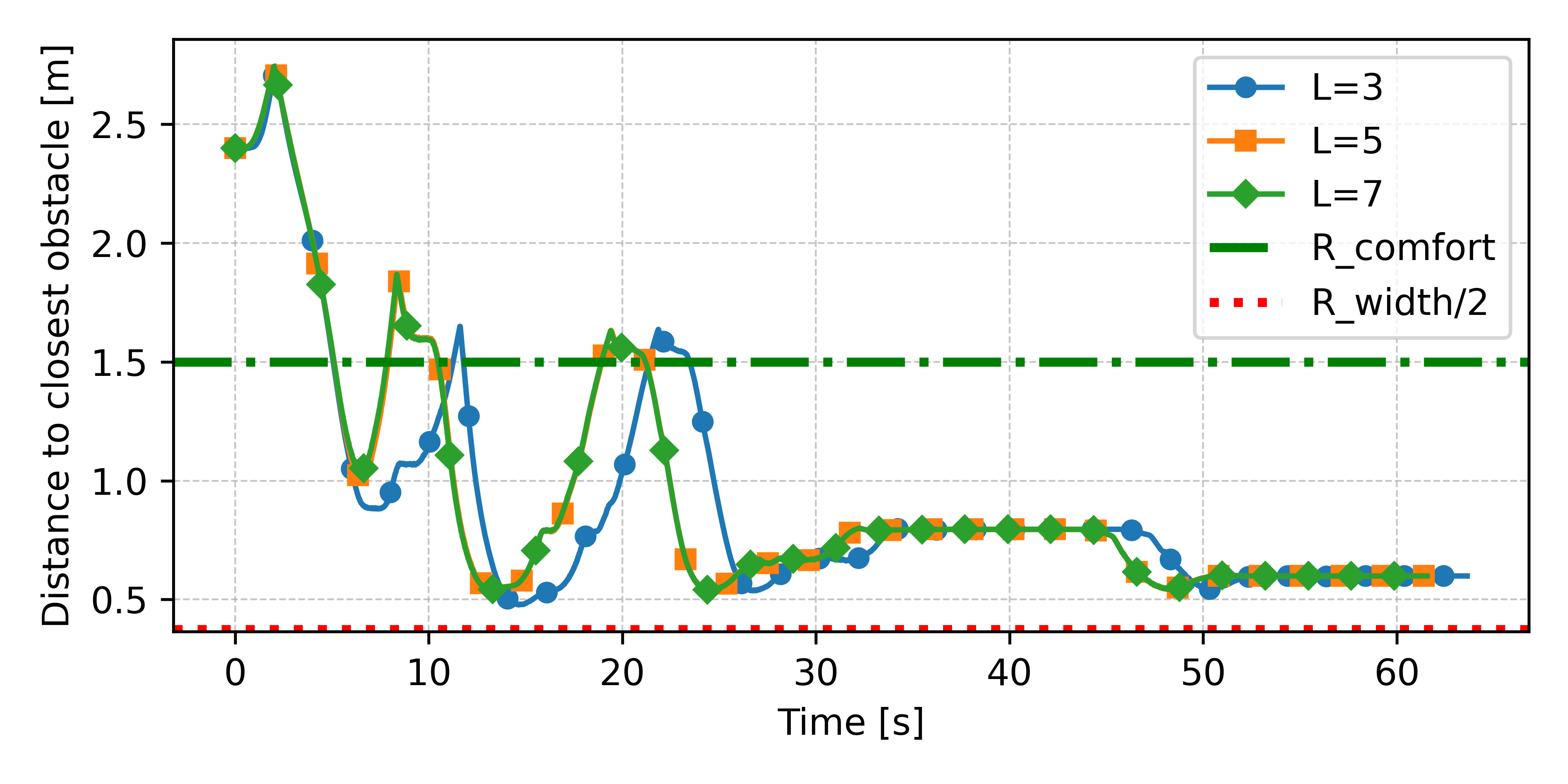}
    \vspace*{-7mm}
    \caption{Without obstacles.}
    \label{fig:dist_to_obj1}
    \end{subfigure}
    
    \begin{subfigure}{1\linewidth}
    \includegraphics[width=1\linewidth]{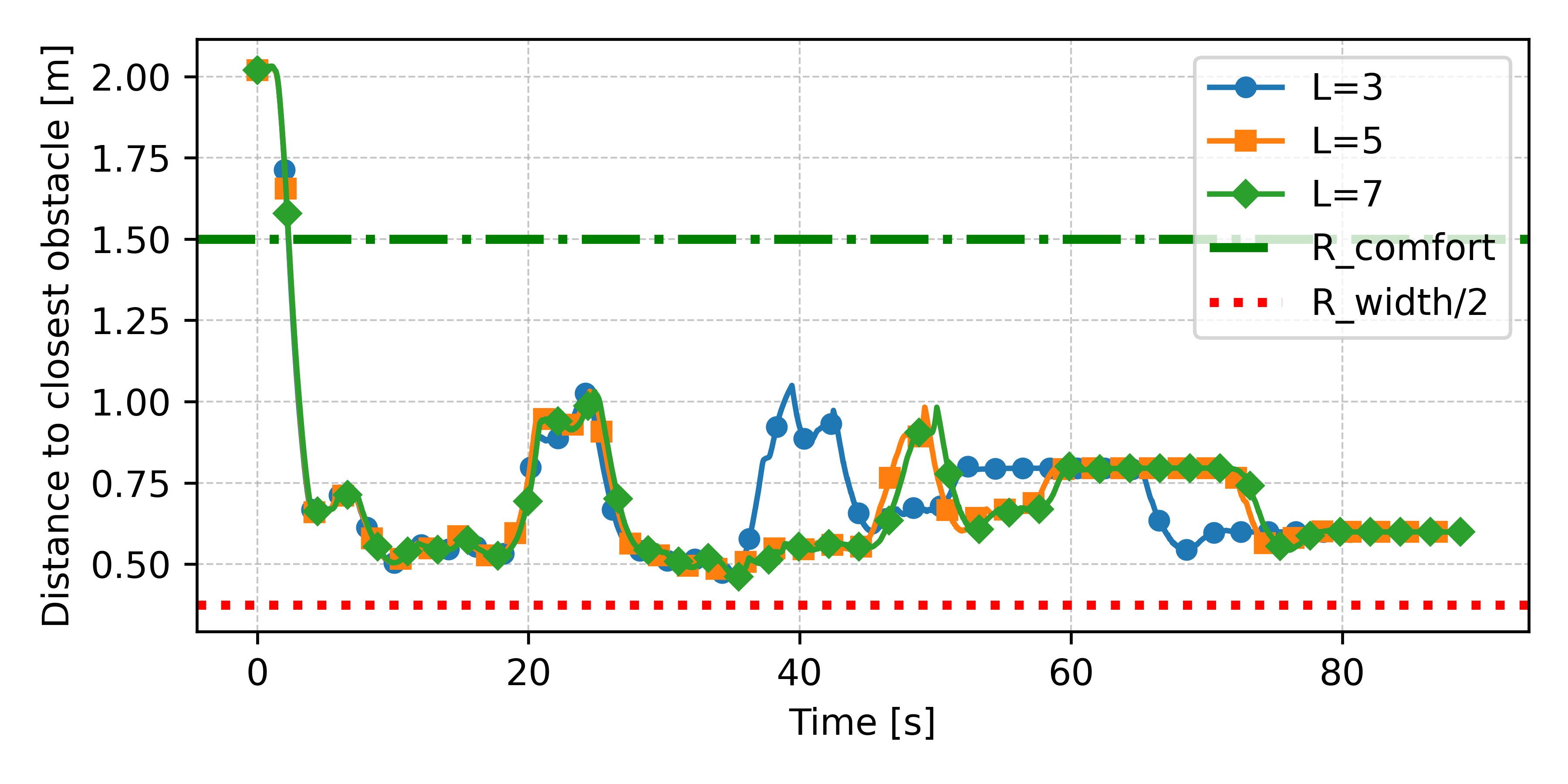}
    \vspace*{-7mm}
    \caption{With obstacles.}
    \label{fig:dist_to_obj2}
    \end{subfigure}
    \caption{Distance to closest obstacle over time.}
    \label{fig:dist_to_obj}
\end{figure}
\begin{figure}[b]
    \centering
    \begin{subfigure}{1\linewidth}
    \includegraphics[width=1\linewidth]{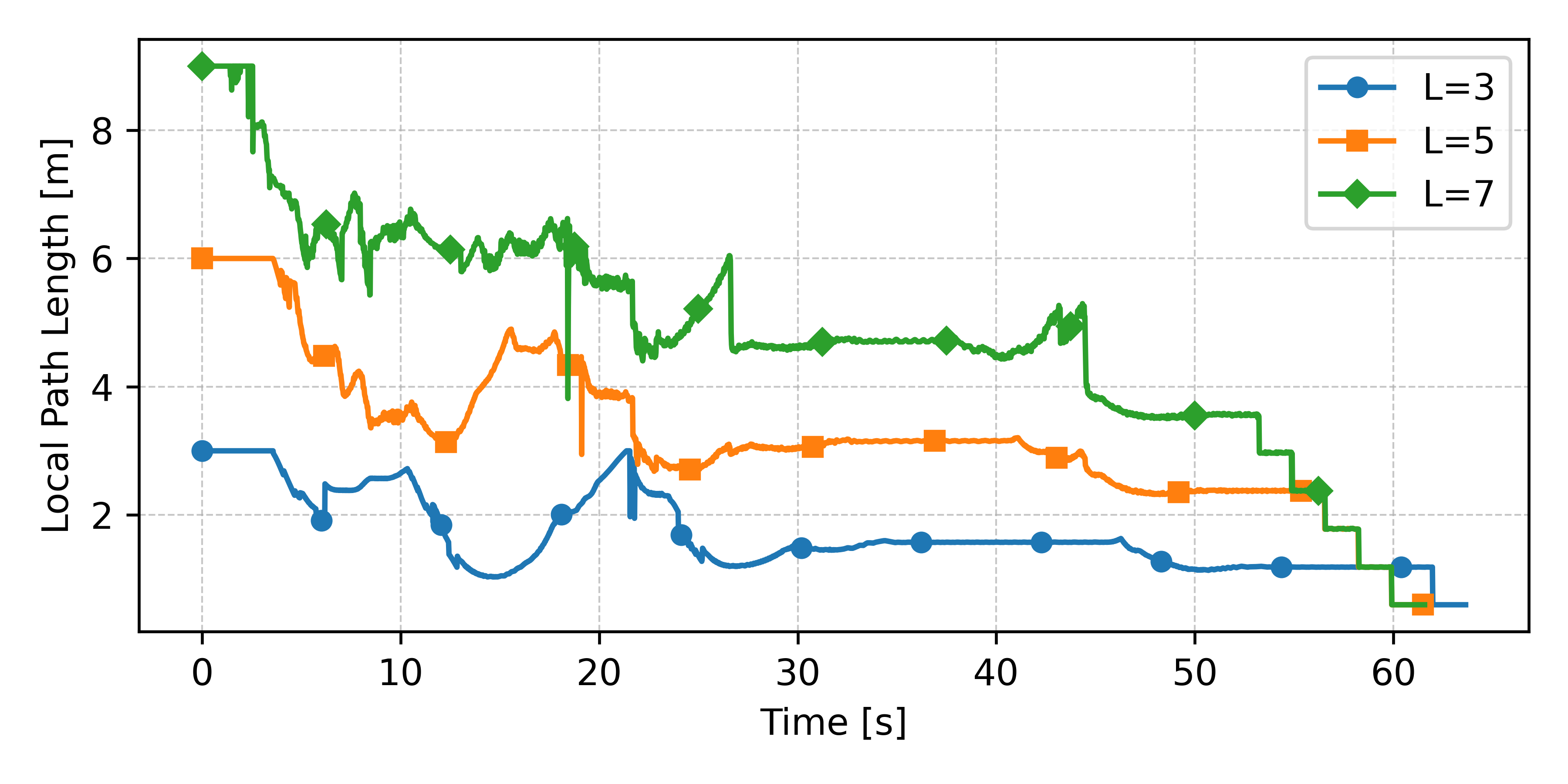}
    \vspace*{-7mm}
    \caption{Without obstacles.}
    \label{fig:local_length1}
    \end{subfigure}

    \begin{subfigure}{1\linewidth}
    \includegraphics[width=1\linewidth]{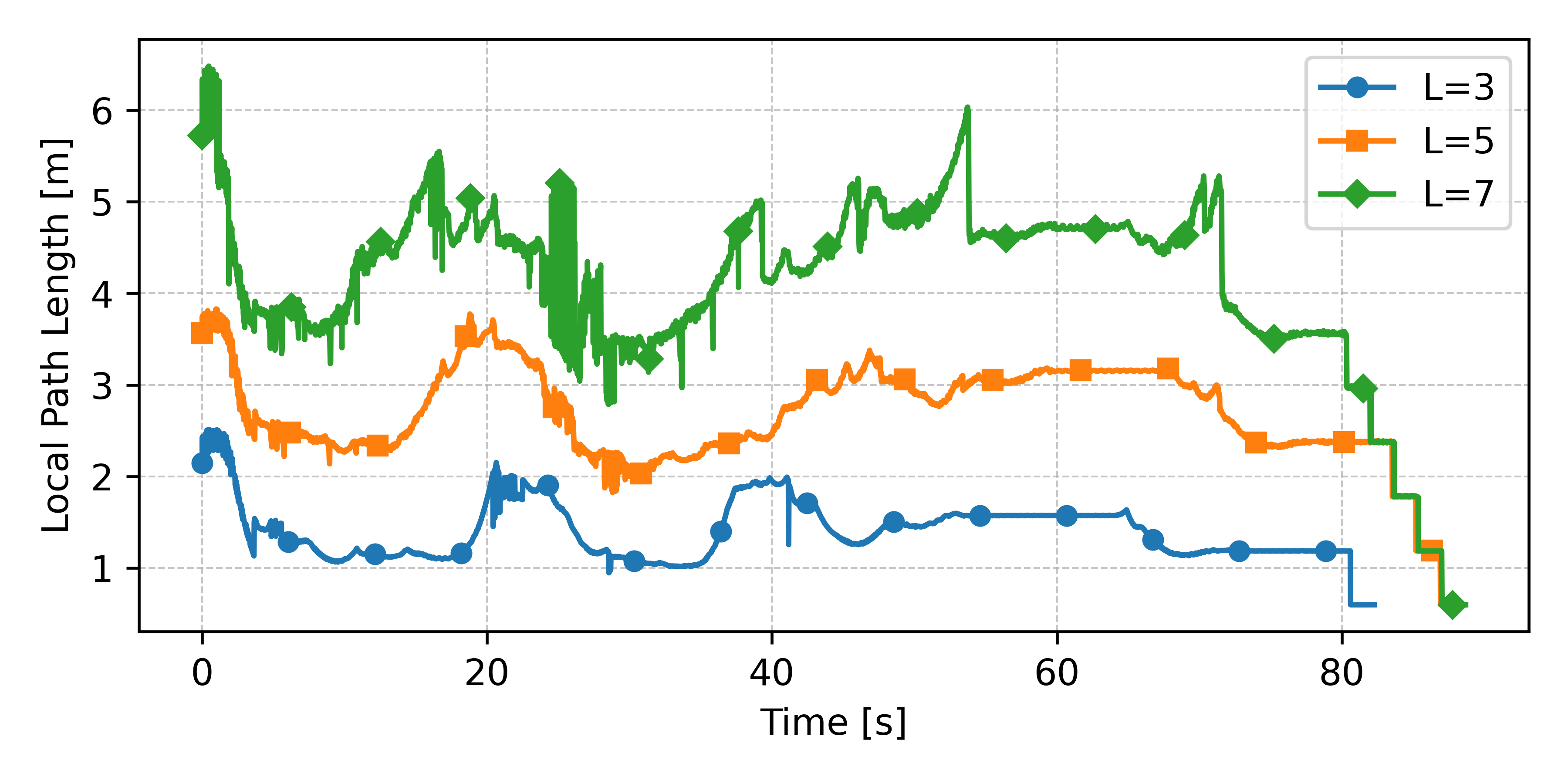}
    \vspace*{-7mm}
    \caption{With obstacles.}
    \label{fig:local_length2}
    \end{subfigure}
    \caption{Length of the local path generated over time.}
    \label{fig:local_length}
\end{figure}
\begin{figure*}[]
    \centering
    \begin{subfigure}{0.95\textwidth}
    \includegraphics[width=\linewidth]{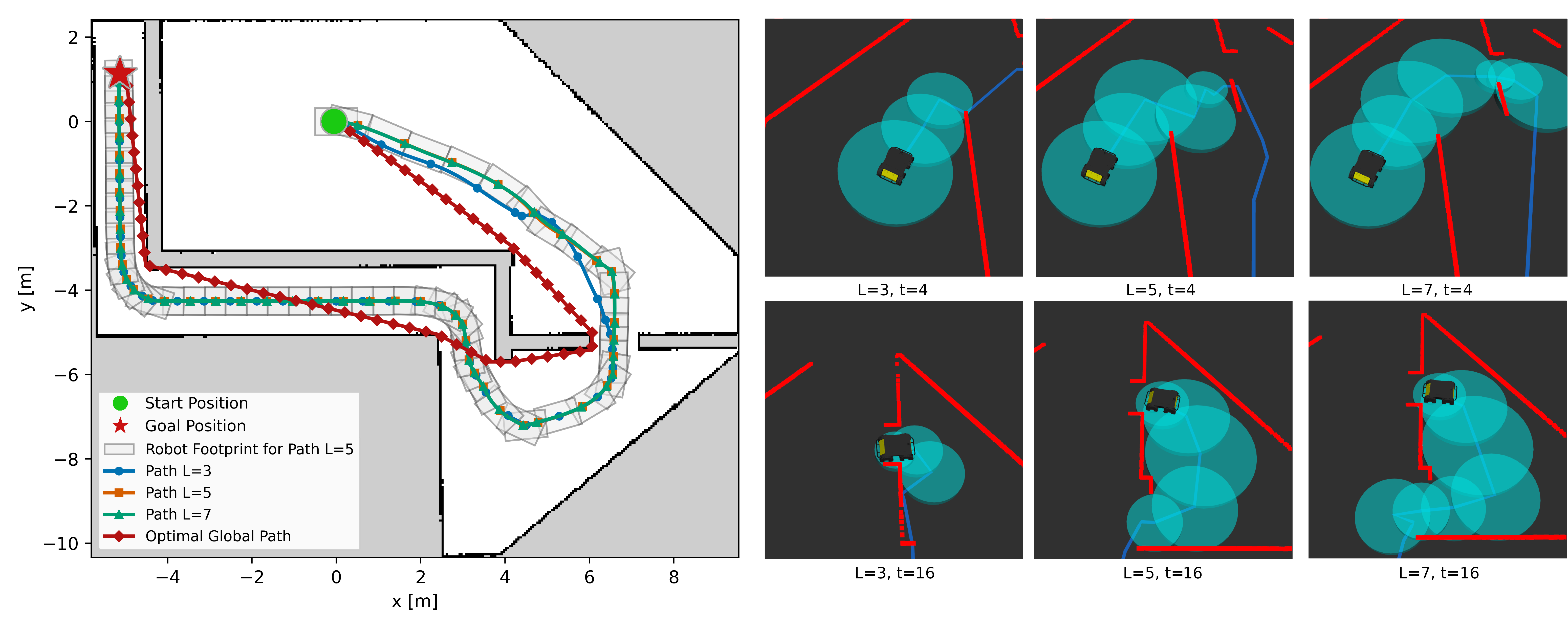}
    \vspace*{-7mm}
    \caption{Environment without obstacles.}
    \label{fig:res_paths1}
    \end{subfigure}

    \begin{subfigure}{0.95\textwidth}
    \includegraphics[width=\linewidth]{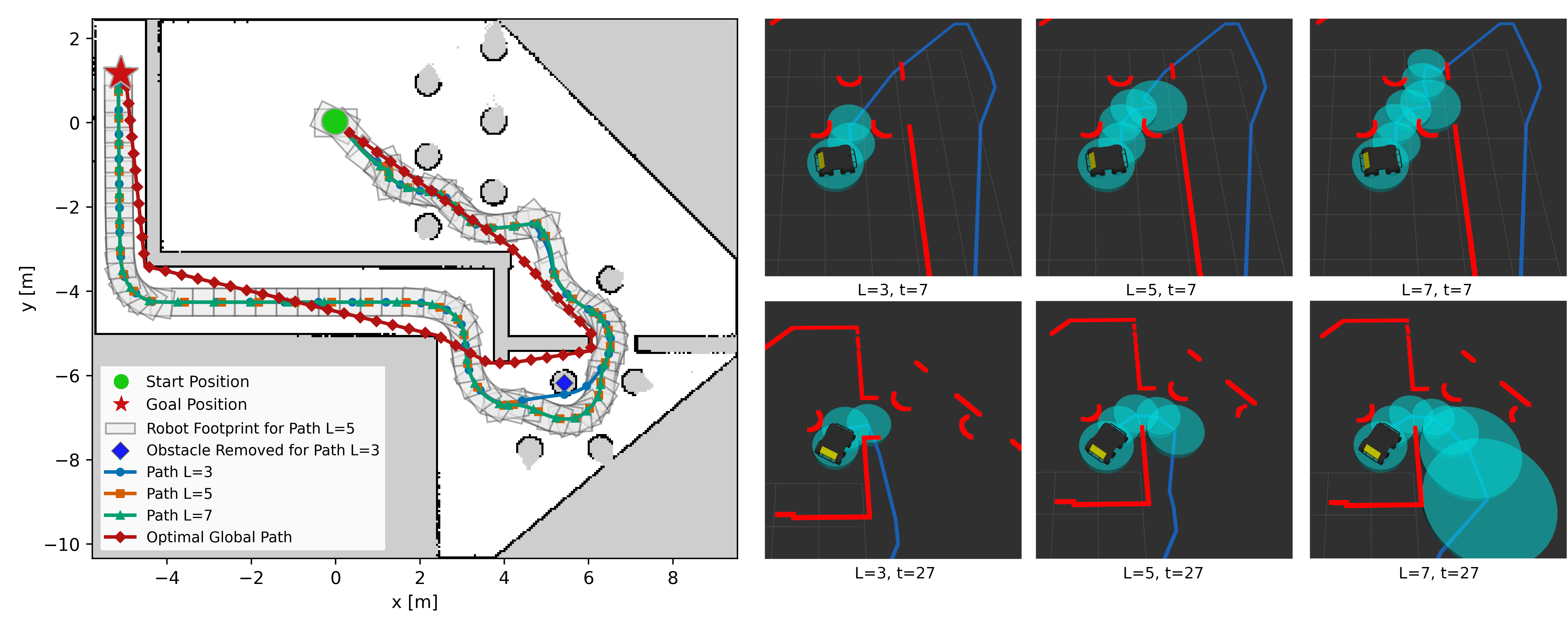}
    \vspace*{-7mm}
    \caption{Environment with obstacles. For $L = 3$, the marked obstacle was removed to obtain a successful run.}
    \label{fig:res_paths2}
    \end{subfigure}
    \caption{Robot paths generated by the proposed method for different configurations of $L$, along with snapshots of the circular regions.}
    \label{fig:res_paths}
\end{figure*}
\begin{table}[h]
    \centering
    \begin{tabular}{|l|l|l|l|}
    \hline
        \multicolumn{4}{|c|}{Environment without obstacles} \\ \hline
        Value of $L$ & 3 & 5 & 7  \\ \hline
        Average Computation Time (ms) & 0.453 & 0.495 & 0.612 \\\hline
        Average Local Path Length (m) & 1.679 & 3.260 & 4.850 \\\hline
        Average Distance to Obstacle (m) & 0.913 & 0.944 & 0.945 \\\hline
        Average Forward Velocity (m/s) & 0.470 & 0.493 & 0.491\\ \hline
        Average Angular Velocity (rad/s) & 0.232 & 0.217 & 0.217\\ \hline
        Path Length (m) & 28.48 & 28.74 & 28.79\\ \hline
        Path Time (s) & 63.64 & 61.55 & 61.57\\ \hline
        \hline
        \multicolumn{4}{|c|}{Environment with obstacles} \\ \hline
        
        Value of $L$ & 3* & 5 & 7  \\ \hline
        Average Computation Time (ms) & 0.295 & 0.424 & 0.586 \\\hline
        Average Local Path Length (m) & 1.399 & 2.671 & 4.127 \\\hline
        Average Distance to Obstacle (m) & 0.720 & 0.699 & 0.697 \\\hline
        Average Forward Velocity (m/s) & 0.374 & 0.358 & 0.360\\ \hline
        Average Angular Velocity (rad/s) & 0.264 & 0.271 & 0.273\\ \hline
        Path Length (m) & 28.21 & 28.95 & 28.96\\ \hline
        Path Time (s) & 82.27 & 88.56 & 88.69\\ \hline
    \end{tabular}
    \caption{List of all the metrics measured from the simulation. * For $L = 3$ an obstacle was removed, resulting in a slightly different path.}
    \label{tab:res}
\end{table}
The method was validated in two Gazebo simulation environments using a Husky robot. The proposed method uses the parameters shown in Table~\ref{tab:vals} in all runs. 
The tests were conducted on a computer with an AMD 5850U CPU running Linux kernel version 6.17.7.
The simulations were performed with 3, 5, and 7 circular regions ($L$).

An optimal global path was generated between a start and goal point in a simulated environment featuring tight openings and narrow corridors. The second environment is identical to the first, except that additional obstacles were introduced, making the local navigation problem more challenging, as shown in Figure~\ref{fig:res_paths}\footnote{Video of simulation results: \\https://www.youtube.com/watch?v=RnOdL5bxskw}. The figure also illustrates the resulting local paths produced using different values of $L$. 

All configurations successfully followed the complete global path in the environment without obstacles, whereas the configuration with $L = 3$ became stuck in a deadlock in the environment with obstacles while attempting to navigate around the obstacle highlighted in Figure~\ref{fig:res_paths2}. To obtain a complete path for the case $L = 3$, the obstacles were removed for that particular run.

During the simulation, the computation time of the local planner was measured (Figure~\ref{fig:res_comput_time}), along with the length of the local path (Figure~\ref{fig:local_length}), distance to the closest obstacle (Figure~\ref{fig:dist_to_obj}), the forward and angular velocities, and the final travel distance and time. 
Table~\ref{tab:res} lists all measured metrics from the simulation.

\section{Discussion}
As shown in Section~\ref{sec:val}, the proposed Fast Expanding Safe Circular Regions (FESCR), in combination with a simple velocity controller, results in a computationally efficient local planner. The simulation results indicate that the method can handle what may be considered a difficult static environment, with many narrow openings, corridors, and obstacles. Using the average speed and the average local path length, a time horizon for the method can be estimated. This results in horizons of approximately 3.7, 7.5, and 11 seconds, for $L=3$, $L=5$, and $L=7$, respectively, in the environment with obstacles. In comparison, existing MPC-based methods \cite{Brito2019,Liu2024} typically employ a prediction horizon of 1 to 5 seconds while requiring significantly more computation.

The simulations also highlight that a long prediction horizon can be important in certain scenarios. For example, when using three circular regions ($L = 3$), the planner is unable to find a feasible local path around the obstacle marked in Fig.~\ref{fig:res_paths2}. This scenario is particularly challenging because the global path lies between a wall and an obstacle, preventing the robot from following it directly. Instead, the robot must travel around the opposite side of the obstacle, which requires a sufficiently long horizon so that the local planner can reach a configuration that enables bypassing the obstacle.

It is important to note that, compared to existing methods, the length of the local path generated by the proposed approach varies with the environment. As shown in Table~\ref{tab:res}, the environment without obstacles results in a longer average local path for all configurations compared to the environment with obstacles.
Depending on the number and arrangement of obstacles, the local path length lies between $(L - 1) \cdot R_\text{width}/2$ and $(L - 1)\cdot R_\text{comfort}$ meters. As also shown in the results in Section~\ref{sec:val}, there is barely any difference between the paths generated with $L = 5$ and $L = 7$. This indicates diminishing returns when increasing the number of circular regions beyond a certain point. This effect is expected because the proposed method relies solely on the current point cloud; thus, as $L$ increases, the predicted path eventually reaches the boundary of the point-cloud sensing range and begins extending into empty space. 

Another important property of FESCR is its attempt to maintain the $R_\text{comfort}$ distance to all obstacles. As shown in Figure~\ref{fig:dist_to_obj1}, the robot prioritizes keeping the $R_\text{comfort}$ distance whenever possible. This results in a safer path, although at the cost of a longer travel distance. By reducing $R_\text{comfort}$, the robot would achieve shorter total travel distances, but at the expense of reduced safety margins. 

However, in tight spaces, such as the corridors in the second half of the paths shown in Figure~\ref{fig:res_paths}, the robot will stay near the center of the corridor instead of strictly following the global path that runs close to the wall.

Overall, the results suggest that $L = 5$ provides the best balance for the proposed method, offering strong performance while successfully navigating the environment without requiring the removal of obstacles.

\section{Conclusion and Future Work}
This article introduces the concept of Fast Expanding Safe Circular Regions (FESCR) combined with a simple velocity controller for local path planning. In simulation, the proposed method is shown to handle complex environments effectively while remaining computationally efficient.

Future work will focus on expanding the types of controllers used with FESCR, such as model predictive control and machine learning approaches. Additional directions include supporting more types of robots, such as robots with Ackermann steering.

\bibliography{ifacconf}             

\end{document}